\pdfoutput=1

\documentclass[11pt]{article}
\usepackage{multirow}
\usepackage[table,xcdraw]{xcolor}
\usepackage{graphicx}
\usepackage{enumitem}
\usepackage{footnote}
\makesavenoteenv{tabular}
\usepackage{array}

\usepackage[table]{xcolor}
\usepackage{amsmath}
\usepackage{hyperref}

\usepackage[preprint]{acl}

\usepackage{times}
\usepackage{latexsym}
\usepackage{subcaption}
\usepackage[normalem]{ulem}
\usepackage[T1]{fontenc}

\usepackage[most]{tcolorbox}

\tcbset{
  examplebox/.style={
    colback=white,
    colframe=black,
    boxrule=0.5pt,
    arc=0mm,
    left=2mm,
    right=2mm,
    top=1mm,
    bottom=1mm,
    enhanced,
    breakable,
    sharp corners,
    before skip=6pt,
    after skip=6pt
  }
}

\usepackage[utf8]{inputenc}

\usepackage{microtype}
\usepackage{booktabs}
\usepackage{tabularx}
\usepackage{inconsolata}
\title{Explanation Bias is a \textit{Product}: Revealing the Hidden Lexical and \\Position Preferences in Post-Hoc Feature Attribution}

\author{Jonathan Kamp \\
    Computational Linguistics and Text Mining Lab, Vrije Universiteit Amsterdam \\
    \\
    \texttt{j.b.kamp@vu.nl}}

\author{Jonathan Kamp$^{1}$\hspace{1cm}Roos Bakker$^{2,3}$\hspace{1cm}Dominique Blok$^{2}$ \\
    $^{1}$Computational Linguistics and Text Mining Lab, Vrije Universiteit Amsterdam \\
    $^{2}$TNO--The Netherlands Organization for Applied Scientific Research, The Hague \\
    $^{3}$Leiden University Centre for Linguistics (LUCL), Leiden University, Leiden \\
    \texttt{j.b.kamp@vu.nl, \{roos.bakker, dominique.blok\}@tno.nl}}

\begin{document}
\maketitle
\begin{abstract}
Good quality explanations strengthen the understanding of language models and data. Feature attribution methods, such as Integrated Gradient, are a type of post-hoc explainer that can provide token-level insights. However, explanations on the same input may vary greatly due to underlying biases of different methods. Users may be aware of this issue and mistrust their utility, while unaware users may trust them inadequately. In this work, we delve beyond the superficial inconsistencies between attribution methods, structuring their biases through a model- and method-agnostic framework of three evaluation metrics. We systematically assess both lexical and position bias (\textit{what} and \textit{where} in the input) for two transformers; first, in a controlled, pseudo-random classification task on artificial data; then, in a semi-controlled causal relation detection task on natural data. We find a trade-off between lexical and position biases in our model comparison, with models that score high on one type score low on the other. We also find signs that anomalous explanations are more likely to be biased.
\end{abstract}

\section{Introduction}

\begin{figure}[th]
    \centering
    \includegraphics[width=0.48\textwidth]{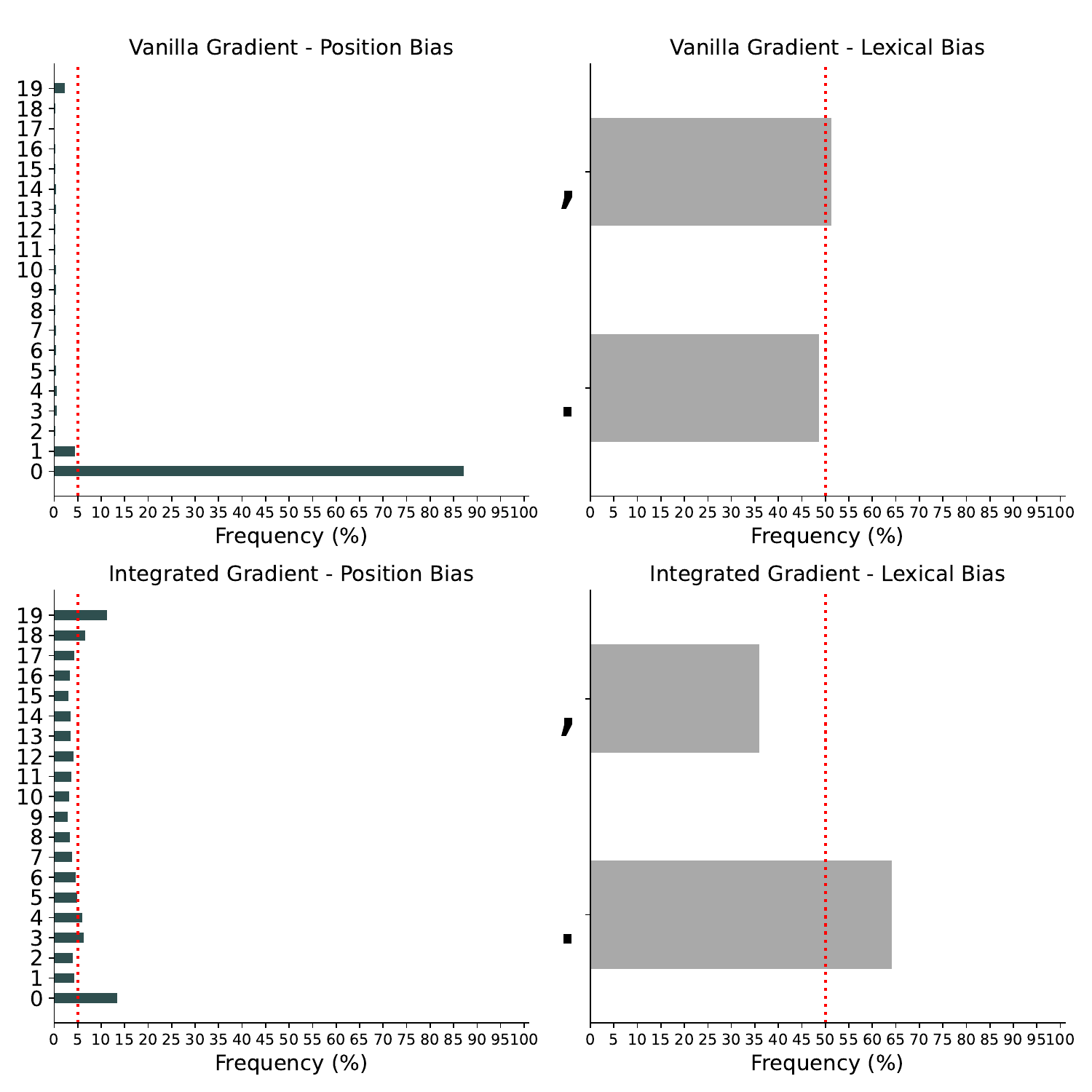}
    \caption{Bias distributions of top-1 token explanations from \texttt{ModernBERT} classifiers over \textit{positions} (left; 0--19) and \textit{lexical elements} (right; `\textit{,}' and `\textit{.}') in the input. Here we compare two feature attribution methods: Vanilla Gradient (top); Integrated Gradient (bottom). The dotted line indicates the position where we would expect the predictions to be in the absence of bias.}
    \label{fig:figure1}
\end{figure}

Feature attribution is a post-hoc explanation technique that aims to determine which input tokens contributed most to a language model's prediction. Different attribution methods have been proposed, but they tend to exhibit inconsistent token explanations on the same input and given the same backbone model \citep{neely2022song}. A major problem that comes with attribution inconsistency is the lack of adequate domain knowledge: methods might either be adopted without considering the risks of unfaithful explanations (i.e.\ explanations that do not reflect the true decision mechanisms of the model), or might be coarsely deemed unreliable and discarded altogether. Their inconsistent explanation behaviour can manifest itself through various symptoms that reflect some underlying systematic deviation, i.e.\ a \textit{bias}, that resides in or surfaces through the attribution method. 
For example, \citet{lyu-etal-2024-towards} argue that, in cases where content words (e.g. \textit{great}) are expected to be more informative than function words (e.g. \textit{the}), methods that assign higher importance to function words raise the question of whether the model is truly relying on them or whether the method fails to faithfully reflect the model's reliance on content words. We talk about a \textit{lexical bias} if the attribution is systematically assigned to specific lexical elements.  

However, lexical information might only be a part of the full story, as \textit{positional} information likely plays a role in bias, too. Figure \ref{fig:figure1} illustrates the bias distributions for token position and token type of two explanation methods on the same classification task on artificial sequences of periods and commas. Vanilla Gradient \citep{simonyan2014deep} is very biased towards the start of the input texts (index 0). In contrast, the more elaborate Integrated Gradient \citep{sundararajan2017axiomatic} seems more balanced towards position but shows some lexical bias by preferring `.' over `,'.\footnote{We do not only consider words but any element in the lexicon to be lexical elements, including punctuation.}

In this paper, we study the sources of bias. We argue that lexical and position biases can be a product of the \uline{attribution method}, the \uline{model type}, and the \uline{random seed}. In this light, we propose an evaluation framework to investigate the biases exhibited by six different attribution methods combined with a traditional and a recent BERT-based classifier. We assess how these transformers, which differ in pre-training, architectural properties and size, suffer from bias. This allows us to address questions such as: is the modern BERT version less affected? And to what extent are position and lexical bias related? By systematically structuring bias symptoms, we aim to better understand the underlying causes of method--method inconsistencies (also known as \textit{disagreement}) to help users of attribution methods make more informed decisions. We first experiment in a controlled setting with artificial data, followed by a semi-controlled setting on English causal relation detection.

Among our main findings, we show that these model attributions indeed systematically exhibit certain types of bias. Importantly, a \textit{more modern} model does not necessarily entail lower bias, and the proneness for a model to exhibit position bias appears to be unrelated to lexical bias, with scenarios in which one increases, the other decreases. Shared results by both artificial and natural settings show that the attribution method producing the most diverging explanations is also the most biased, and that position bias mostly arises towards the end of the inputs in the traditional BERT model.\footnote{We publicly release our code at\\ \url{https://github.com/jbkamp/repo-PosLex-Bias}.}

\section{Related Work}

Attribution methods are inconsistent compared to one another \citep{krishna2022disagreement, neely2022song, kamp-etal-2023-dynamic}, which makes it challenging to interpret a backbone model. Possible underlying biases, such as the preference for certain lexical classes and their position in the input, may not only affect model understanding but also the cognitive processing of the presented explanations. For example, \citet{jacovi-etal-2023-neighboring} carry out a user study showing that lexical and positional information of highlighted tokens are indicative of the way humans value explanations as important.

Lexical bias in attribution methods has been explored in a handful of tasks, such as sentiment analysis and deception detection \citep{lai2019many}, and reading comprehension-based question answering \citep{ramnath2020towards}. \citet{kamp-etal-2024-role} investigate the disagreement among attribution methods by mapping explanations to lexical classes. They find, on a natural language inference task, that methods may have different biases towards e.g.\ nouns, determiners and punctuation but often still converge to the same syntactic phrase. They speculate that position bias might be induced by the proximity to an important token. The problem is related to the work of \citet{ko-etal-2020-look} and \citet{ben2024impact}, who show that the position of signal words in the input induces a model bias towards a specific target class. \citet{xiao2023efficient} find strong attention scores (\textit{sinks}), on initial tokens LLM inputs, irrespective of their semantics.
\citet{eshuijs-etal-2025-short} instead find that the first most informative word in a span receives the most attention in decoder-only models, hence overruling the weight of the word's position in the span. 

The fact that biases are observed on specific tasks raises the question to what extent biases are task- and data-dependent; some contrasting findings by \citet{attanasio-etal-2022-benchmarking} versus \citet{atanasova2020diagnostic} seem to indicate a task-dependent component in the source of disagreement. We differ from previous work in that we assess both lexical and position bias in a controlled setting to isolate the findings from task-related aspects. To our knowledge, we are the first to do so.

Two factors that may directly affect feature attribution are the size of the fine-tuning data and the size of the model (i.e. the number of parameters). \citet{zhou-etal-2024-explaining} explore the effect of fine-tuning data size, showing that more data usually leads to greater \textit{plausibility} (i.e.\ alignment with human rationales) but not to greater \textit{faithfulness} (i.e.\ alignment with model internals). 
Notably, they do not find a clear effect of model size (Vicuna, 13B \citep{vicuna2023} vs. RoBERTa-large, 355M \citep{liu2019roberta}) on plausibility or faithfulness; instead, the results strongly depend on the dataset and the attribution method. Conversely, \citet{heyen2024the} compare LIME \citep{ribeiro-etal-2016-trust} on four versions of DeBERTaV3 \citep{hedebertav3} (22--304M parameters), and find that explanations do not become more plausible in larger models, but do become more faithful. The caveat is that metrics that assess faithfulness can be unreliable, too. in the same study, \citet{heyen2024the} find that \textit{comprehensiveness}, which approximates faithfulness as the prediction drop after removing explanation tokens from the input, is not in line with other metrics capturing greater faithfulness for the larger model; this suggests a possible metric limitation. 

We experiment with two BERT-based models. In the process, we also validate two faithfulness metrics: \textit{sufficiency} and \textit{comprehensiveness}. The design of the experiments helps us to explore bias in different data constructions with different levels of control; we begin with an artificial setting in §\ref{sec:experiments_on_artificial_data}, followed by a causal relation detection task in §\ref{sec:experiments_on_causal_relation_data}.

\section{Experiments on Artificial Data}\label{sec:experiments_on_artificial_data}
To study attribution bias, we use the frequency in which token explanations are located at certain input indices (position bias) and belong to a certain word type (lexical bias). We craft the tasks in a controlled setting in such a way that this information should not play a role in solving the task (which should not be solvable). 
If attributions are biased, they should manifest this bias consistently across runs and datasets. In §\ref{sec:experiments_on_causal_relation_data}, we complement this randomised setting with a natural language task.

\subsection{Data} 
We craft three artificial datasets for a binary classification task. For each, we generate a set of 8,000 instances for training, 1,000 for validation and 1,000 for testing. Each instance of fixed length 20 (an average sentence length) is populated with different lexical elements. Within our framework, a set of lexical elements has been chosen that can easily be replaced. The idea is that position bias arises irrespective of the lexical information. Below, we describe these three artificial datasets:

\paragraph{\texttt{noun-det-period}} The chosen lexical classes follow \citet{kamp-etal-2024-role}, whose initial insights show preferences for specific parts-of-speech. We include a frequent noun (`table') and determiner (`the'), and a period (`.'). Each token can occur zero or more times. Token positions in the sentences are allocated randomly. Class labels are random and equally distributed.

\paragraph{\texttt{period-comma}} Each instance is a random sequence of `.' and `,' tokens, each of which can occur 0+ times. The absence of both content and function words may highlight the lexical bias in certain attribution methods that favour punctuation. Class labels are random and equally distributed. 

\paragraph{\texttt{unique-punctuation}} Each instance is a unique, random ordering of the set of unique punctuation marks \{`.', `,', `;', `:', `!', `?', `-', `\_', `(', `)', `[', `]', `\{', `\}', `/', `*', `\#', {``}'{''}, `"', {``}`{''}\}. The uniqueness of these lexical elements should facilitate faithfulness assessment (introduced in §\ref{sec:bias_metrics}) by enabling us to measure the weight of a specific token in the input. Class labels are random and equally distributed. 

\subsection{Models}
We compare two transformer models with a sequence classification head, namely \texttt{BERT}: bert-base-uncased, 110M \citep{devlin-etal-2019-bert} and \texttt{ModernBERT}: ModernBERT-base, 149M \citep{modernbert}. We investigate whether a more recent BERT architecture suffers less from bias. After introducing the bias metrics in §\ref{sec:bias_metrics}, we speculate further on how model properties may relate to bias.

\subsection{Explanations} 
We compute token attribution scores for each of the input sequences. We ignore start-of-sequence and end-of-sequence tokens, such as [CLS] and [SEP], after computing the attributions in order to focus the analysis on the original 20 tokens per input. We test six different attribution methods: Partition SHAP, or \textbf{PartSHAP} \citep{lundberg2017unified} \textbf{LIME} \citep{ribeiro-etal-2016-trust}; Vanilla Gradient, or \textbf{VanGrad} \citep{simonyan2014deep}, and Integrated Gradient, or \textbf{IntGrad} \citep{sundararajan2017axiomatic}; the latter two multiplied by the input -- \textbf{Grad\,×\,I} and \textbf{IntGrad\,×\,I} \cite{shrikumar2017learning}. We select the top-$k$ attributions per instance and set $k=1$, i.e.\ the single strongest attribution preference. This choice ensures a fair comparison with the real world task in §\ref{sec:experiments_on_causal_relation_data}, where we expect a single word signal to be indicative of the class label.

\subsection{Bias Metrics}\label{sec:bias_metrics}
Given an attribution method, the instance-wise \mbox{top-1} explanations computed on a test dataset form a frequency distribution over the lexical categories and position indices (Figure \ref{fig:figure1}). These distributions may reflect a bias from the method itself. However, possible biases from pre-training and random weight initialisation would also surface in this way. To properly differentiate between bias types, we propose an analysis framework that includes three key metrics for bias quantification: (a) \textbf{Bias-cons}, the consistency of the frequency bias between the different random seeded runs; (b) \textbf{Bias-agg}, the bias of the aggregate distribution (of a specific attribution method) over 10 fine-tuned models trained with different random seeds; (c) \textbf{Bias-attr}, the bias that is specific to the attribution method, compared to other methods. In principle, the framework is applicable to any attribution method and top-$k$ value.

Explanation evaluation can be problematic when the test data is out-of-distribution (OOD) with respect to the training data \citep{hase2021out}. This does not apply to our setting, as we sample the training and test instances from the same data distribution. While the artificial instances might be unlikely or rare in pre-training, the highly frequent tokens that compose them (e.g.\ `.', `,') have regular pre-trained word embeddings and should therefore not be OOD with respect to the pre-training data. Technically, the latter OOD condition could even benefit our setup: as the artificial experiments target feature attribution bias \textit{irrespective} of the data, the modelling of OOD examples would make it less likely to be influenced by pre-training data cues. Still, to rule out possible random noise in the top-1 frequency distributions caused by (improbable) OOD training effects, we report the results of a statistical randomisation test in §\ref{sec:artificial_results}. This test shows that the artificial data experiments produce overall systematic bias patterns that cannot be explained by random noise alone. 

\paragraph{\uline{Bias-cons} for inter-seed comparison}
We investigate to what extent biases are consistent across different random seeds. Since we are interested in the distribution of top-$k$ selections over input tokens, we measure the mean pairwise Jensen-Shannon distance (\textbf{Bias-cons}) between the normalised distributions yielded by different runs. For this, we consider the unique model pairs $(m',m'')$ formed from the set of different runs $\mathcal{M}$ and we average the JS distance over the number of combinations, i.e.\ $\binom{|\mathcal{M}|}{2} = \frac{|\mathcal{M}|(|\mathcal{M}|-1)}{2}$. JS ranges from 0 (least divergent/biased) to 1 (most divergent/biased). The greater the distance, the greater the impact of random initialisation on the explanations. Formally, for each model type $m\in \{\text{\texttt{BERT, ModernBERT}}\}$ and each attribution method $a \in \mathcal{A}$, let $P^{(m')}_{m,a}$ or $P^{(m'')}_{m,a}$ denote the bias distribution (over positions or lexical elements in the input) for a single run $m'$ or $m''$ (i.e., a specific random seed):
\begin{equation}
\begin{aligned}
\text{Bias-cons}_{m,a} = \\ \frac{1}{\frac{|\mathcal{M}|(|\mathcal{M}|-1)}{2}} 
\sum_{\substack{m',\,m'' \in \mathcal{M} \\ m' < m''}} 
\text{JS}\left( P^{(m')}_{m,a} \,\|\, P^{(m'')}_{m,a} \right)
\end{aligned}
\end{equation}

\paragraph{\uline{Bias-agg} for inter-model comparison}
We investigate the extent to which attribution methods are biased in the light of a model-to-model comparison. For each model type $m$, we compute the Jensen-Shannon distance ($\text{JS}$) between each method $a$'s aggregate bias distribution $P_{m,a}$ and a baseline. $P_{m,a}$ is computed by summing token-level bias distributions across all runs $\mathcal{M}$ and normalising to form a probability distribution. The baseline consists of a normalised uniform distribution $U$ over the positions and lexical elements for the specific dataset, and implies the absence of bias. $U$ is valid in a controlled setting and might need adjustment based on the nature of the task. We use a different baseline for causal relation detection in §\ref{sec:experiments_on_causal_relation_data}. We refer to this distance metric as \textbf{Bias-agg}, the bias of the aggregated distributions. \textit{Aggregate} also refers to the fact that this bias conflates both model bias and attribution method bias.
\begin{equation}
    \text{Bias-agg}_{m,a} = \text{JS}\left(P_{m,a} \,\|\, U\right)
\end{equation}

\paragraph{\uline{Bias-attr} for inter-method comparison} If the model were the only source of bias, all attribution methods would exhibit this bias consistently for the same model, same random seed. This is not the case (attributions differ across methods), indicating that at least some attribution methods introduce bias independently from the model. We define attribution bias (\textbf{Bias-attr}) as the averaged pairwise JS distance between the target method $a$ and each other method $a'$ sampled from $\mathcal{A}$.
\begin{equation}
    \text{Bias-attr}_{m,a} = \frac{1}{|\mathcal{A}\!\setminus\! \{a\}|} \sum_{a' \in \mathcal{A} \setminus \{a\}} \text{JS}(P_{m,a} \| P_{m,a'})
\end{equation}

\noindent It is unknown which of the attribution methods in $\mathcal{A}$ are the most biased. Based on the principle of reproducibility in science \citep{popper2005logic}, a finding that is confirmed through different means is more likely to be correct: can a majority consensus among methods be indicative of faithfulness? Are outlier explanations more likely to be wrong? To answer these open questions, we investigate whether Bias-attr and Bias-agg behave similarly. In other words, are attribution methods that differ mostly from the rest (high Bias-attr) or are most similar to the rest (low Bias-attr) respectively the most and least biased with respect to a baseline expressing absence of bias (high and low Bias-agg)?

\paragraph{Faithfulness metrics: a sanity check} Determining the faithfulness of attributions helps us understand whether the explanations reflect the real feature contributions to the model's prediction. In our controlled setting, a method with high Bias-agg should also be relatively unfaithful, as the model should not have associated any class signals to positions or lexical elements. Do we observe this?

We assess faithfulness through two widely used metrics: \textbf{\textit{sufficiency}} (\textit{suff}) \citep{lei2016rationalizing}, which is obtained by removing the top-1 tokens (the rationale $r_i$) from the input $x_i$ and observing the change in probability towards the ground truth label; \textbf{\textit{comprehensiveness}} (\textit{cmp}) \citep{yu2019rethinking}, which is the change in probability obtained by removing the context ($x_i\setminus r_i$) around the top-1 tokens from $x_i$. Formally, following \citet{deyoung2020eraser}:
\begin{equation}
    \text{suff}(D) = \frac{1}{n} \sum_{i=1}^{n} \left( m'(x_i)_j - m'(r_i)_j \right)
\end{equation}
\begin{equation}
    \text{cmp}(D) = \frac{1}{n} \sum_{i=1}^{n} \left( m'(x_i)_j - m'(x_i\setminus r_i)_j \right)
\end{equation}

\noindent where scores are averaged over dataset $D$ and where $m'(\cdot)_{j}$ returns a probability by the fine-tuned model $m'$ for class $j$ on the full instance ($x_i$) and on the corrupted instance ($r_i$ or $x_i\setminus r_i$). 

This sanity check assesses the utility of these metrics on poorly performing models, currently an underresearched scenario. We expect the detection of feature faithfulness to be challenging when the model itself does not rely on specific signals.

\paragraph{How will model type relate to bias?} Most of the time, expecting a model to \textit{do better} merely based on its size is not enough. To speculate more precisely how our models may relate to bias, we take into account the following three considerations. (i) \citet{ben2024impact} directly relate a form of position bias to the type of positional embeddings in the model. Contrary to \texttt{BERT}, which learnt fixed positional embeddings, \texttt{ModernBERT} uses Rotary Positional Embeddings (RoPE): along with the larger model size, this may lead to a reduced position bias. (ii) Lexical bias may also be reduced by RoPE in \texttt{ModernBERT}, giving more weight to the relative position of a lexical element in the context of the sentence, hence affecting its meaning. (iii) The effect on lexical bias caused by the greater diversity of datasets used by \texttt{ModernBERT} in pre-training is not univocal: it may either mitigate or enforce it.

\subsection{Results}\label{sec:artificial_results}
\paragraph{Randomised settings but non-random explanations}
As expected for this fully controlled experimental setting, our fine-tuned models achieve chance-level f1-scores, in the range .33--.55. Repeated explanations on thousands of examples should cancel out any random noise in the \mbox{top-1} explanation distributions while highlighting any bias consistencies. To further rule out that these explanations are random, we perform a statistical randomisation test on these top-1 distributions. 

Specifically, we measure the extent of Bias-cons (i.e.\ the average inter-seed explanation consistency) if explanations were random, and how this would compare to the observed Bias-cons of our experiments. First, we initiate 10 different fully random null models to mimic different model seeds. Each null model produces 1000 randomly sampled \mbox{top-1} frequency distributions over the 20 possible token positions. In theory, these null distributions should all be close to uniform and therefore produce Bias-cons $\approx$ 0 (i.e.\ high consistency). Second, we compute a single Bias-cons score for these 10 null distributions, and repeat this process a thousand times to obtain a distribution of null Bias-cons scores. Third, we compare the observed Bias-cons for each model$\times$method$\times$dataset to this null Bias-cons distribution: the computed one-tailed p-value indicates the probability of the null distribution to include values greater or equal to the observed Bias-cons. The observed Bias-cons is expected to be relatively small but still greater than the null Bias-cons ($\approx 0$) if differently seeded models produce different but consistent bias patterns. For example, an observed seed A may have a strong bias on the first token position, whereas an observed seed B is similarly distributed to A, but with a stronger bias on the last position; on the other hand, the null seeds $\alpha$ and $\beta$ would both be $\approx$ uniformly distributed, but with smaller differences among one another compared to the observed seeds A and B.

The test statistics yield $p<.05$ for each model$\times$method$\times$dataset combination, with average effect sizes of 3.3$\pm$1.6 (1.4--7.6). This indicates that the observed Bias-cons scores are low, but much larger than the null case. Specifically, there is a reproducible pattern across seeds (even if seeds disagree) and the difference across random seeds is not explainable by random noise alone.

In the next paragraphs, we report the results for all bias metrics based on the \mbox{top-1} attributions across random seeds for each model type and attribution method. We present the full visuals of the bias distributions in Appendix \ref{sec:appendix_barplots}.

\begin{figure}[htbp]
  \centering
  \begin{subfigure}{.49\textwidth}
    \includegraphics[width=\textwidth]{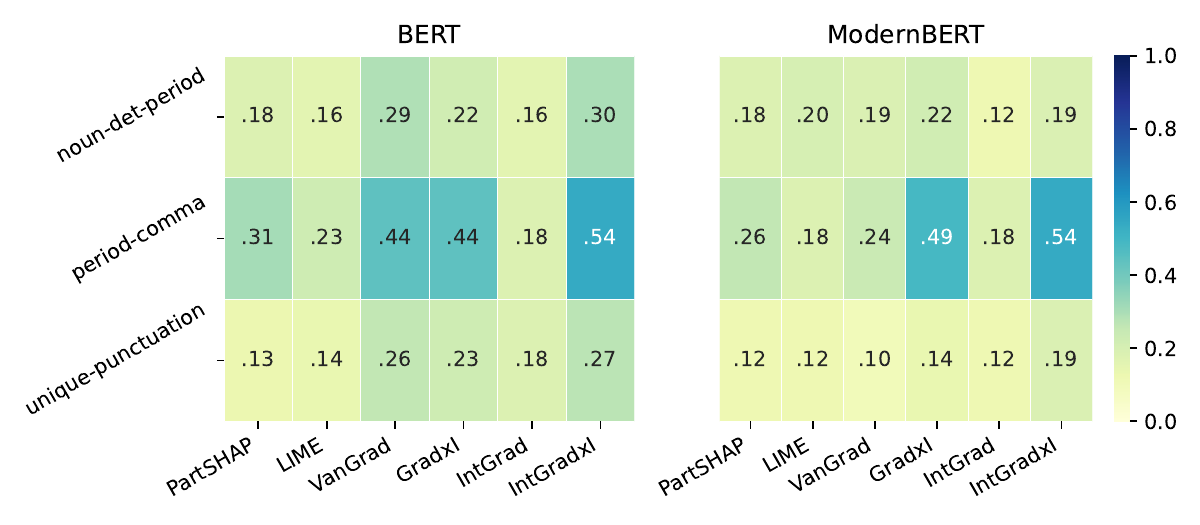}
    \caption{Position Bias-cons.}
    \label{fig:_2plots_indices_JS_interseed}
  \end{subfigure}
  
  \vspace{0.1cm} 
  
  \begin{subfigure}{.49\textwidth}
    \includegraphics[width=\textwidth]{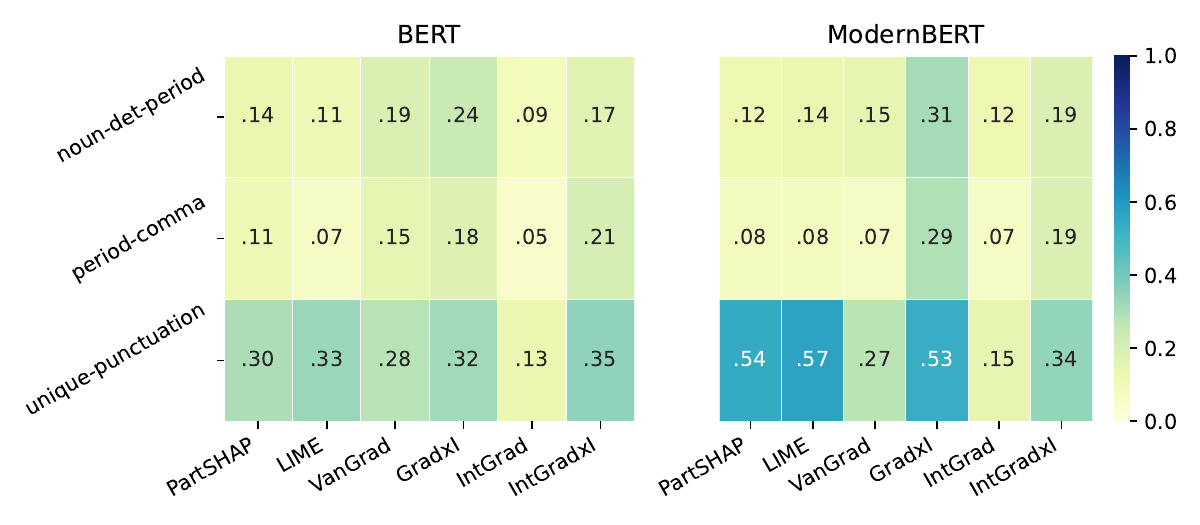}
    \caption{Lexical Bias-cons.}
    \label{fig:_2plots_words_JS_interseed}
  \end{subfigure}

  \caption{Bias-cons: inter-seed consistency on 10 different runs. Low values (light BG) = greater consistency.}
  \label{fig:_2plots_all_JS_interseed}
\end{figure}

\paragraph{Bias-cons} Results are provided in Figure \ref{fig:_2plots_indices_JS_interseed} (position bias) and Figure \ref{fig:_2plots_words_JS_interseed} (lexical bias). Overall, \uline{position bias appears to be more consistent for \texttt{ModernBERT}}, displaying a lower Bias-cons in 12 out of 18 scenarios (3 datasets $\times$ 6 methods) and on average (.21 versus .26). The explanations for \texttt{period-comma} are particularly inconsistent for both model types. Conversely, the \uline{lexical bias is most consistent for \texttt{BERT}} in 11/18 cases and on average (.19 versus .23), with the most visible difference on \texttt{unique-punctuation}.

\begin{table*}[t]
\scriptsize 
\centering
\begin{tabularx}{\textwidth}{cl
>{\centering\arraybackslash}X>{\centering\arraybackslash}X|
>{\centering\arraybackslash}X>{\centering\arraybackslash}X|
>{\centering\arraybackslash}X>{\centering\arraybackslash}X||
>{\centering\arraybackslash}X|>{\centering\arraybackslash}X|
>{\centering\arraybackslash}X|>{\centering\arraybackslash}X|
>{\centering\arraybackslash}X|>{\centering\arraybackslash}X}
\toprule
\multirow{2}{*}{} & \multirow{2}{*}{\textbf{Method}} 
& \multicolumn{2}{c}{\texttt{noun-det-p}} 
& \multicolumn{2}{c}{\texttt{p-comma}} 
& \multicolumn{2}{c||}{\texttt{uniq-punct}} 
& \multicolumn{2}{c}{\texttt{noun-det-p}} 
& \multicolumn{2}{c}{\texttt{p-comma}} 
& \multicolumn{2}{c}{\texttt{uniq-punct}} \\
& & \texttt{B} & \texttt{ModB} & \texttt{B} & \texttt{ModB} & \texttt{B} & \texttt{ModB}
  & \texttt{B} & \texttt{ModB} & \texttt{B} & \texttt{ModB} & \texttt{B} & \texttt{ModB} \\
\midrule
\multirow{7}{*}{\rotatebox[origin=c]{90}{\textit{Position Bias}}}
& PartSHAP & .19 & \cellcolor{blue!15}.22 & \cellcolor{blue!5}.26 & \cellcolor{blue!5}.26 & \cellcolor{blue!15}.23 & .10
              & .13 & .15 & .21 & .26 & .13 & \cellcolor{red!15}.09* \\
& LIME       & \cellcolor{blue!15}.15 & .12 & \cellcolor{blue!15}.23 & .07 & \cellcolor{blue!15}.24 & .03
              & .13 & .11 & .20 & .31 & .11 & .06 \\
& VanGrad    & \cellcolor{blue!15}.39 & .24 & .53 & \cellcolor{blue!15}.66 & \cellcolor{blue!15}.44 & .05
              & \cellcolor{red!15}.27* & \cellcolor{red!15}.19* & \cellcolor{red!15}.38* & \cellcolor{red!15}.51* & \cellcolor{red!15}.23* & .08 \\
& Grad\,×\,I     & \cellcolor{blue!15}.18 & .11 & \cellcolor{blue!15}.37 & .36 & \cellcolor{blue!15}.30 & .05
              & .13 & .11 & .23 & .26 & .15 & .08 \\
& IntGrad    & \cellcolor{blue!15}.15 & .07 & .11 & \cellcolor{blue!15}.16 & \cellcolor{blue!15}23 & .04
              & .16 & .11 & .28 & .26 & .13 & .06 \\
& IntGrad\,×\,I  & \cellcolor{blue!15}.17 & .11 & .31 & \cellcolor{blue!15}.34 & \cellcolor{blue!15}.31 & .09
              & .15 & .11 & .26 & .27 & .14 & .08 \\
\cmidrule(lr){2-14}
& avg.       & \cellcolor{blue!15}.21 & .15 & \cellcolor{blue!5}.31 & \cellcolor{blue!5}.31 & \cellcolor{blue!15}.30 & .06
              & -- & -- & -- & -- & -- & -- \\
\midrule
\midrule
\multirow{7}{*}{\rotatebox[origin=c]{90}{\textit{Lexical Bias}}}
& PartSHAP & \cellcolor{blue!15}.09 & .03 & .02 & \cellcolor{blue!15}.05 & .19 & \cellcolor{blue!15}.22
              & .11 & .20 & .02 & .05 & .19 & .30 \\
& LIME       & .03 & \cellcolor{blue!15}.05 & \cellcolor{blue!5}.03 & \cellcolor{blue!5}.03 & .20 & \cellcolor{blue!15}.23
              & .06 & .16 & .03 & .03 & .19 & .32 \\
& VanGrad    & .04 & \cellcolor{blue!15}.30 & \cellcolor{blue!15}.06 & .01 & .34 & \cellcolor{blue!15}.56
              & .07 & \cellcolor{red!15}.26* & .02 & .07 & \cellcolor{red!15}.31* & \cellcolor{red!15}.54* \\
& Grad\,×\,I     & .02 & \cellcolor{blue!15}.16 & \cellcolor{blue!15}.06 & .04 & .16 & \cellcolor{blue!15}.34
              & .06 & .14 & .02 & \cellcolor{red!15}.09 & .20 & .36 \\
& IntGrad    & .03 & \cellcolor{blue!15}.14 & .02 & \cellcolor{blue!15}.10 & .07 & \cellcolor{blue!15}.18
              & .07 & .13 & .03 & .07 & .19 & .28 \\
& IntGrad\,×\,I  & .12 & \cellcolor{blue!15}.21 & .04 & \cellcolor{blue!15}.11 & .19 & \cellcolor{blue!15}.21
              & \cellcolor{red!15}.14* & .17 & \cellcolor{red!15}.07 & .08 & .26 & .30 \\
\cmidrule(lr){2-14}
& avg.       & .05 & \cellcolor{blue!15}.15 & .04 & \cellcolor{blue!15}.06 & .19 & \cellcolor{blue!15}.29
              & -- & -- & -- & -- & -- & -- \\
\bottomrule
\end{tabularx}
\caption{Position and lexical bias comparison between \texttt{BERT (B)} and \texttt{ModernBERT (ModB)} across datasets. \begin{tabular}{@{}l@{}}\cellcolor{blue!15}Left, Bias-agg\end{tabular}: per dataset, dark background indicates the \textit{model} with highest bias (\textbf{row-wise}), including model avg. \\ \begin{tabular}{@{}l@{}}\cellcolor{red!15}Right, Bias-attr\end{tabular}: per dataset, dark background indicates the \textit{method} with the highest bias (\textbf{column-wise}). (*) indicates \textit{high} Bias-attr methods (right) that correspond to \textit{high} Bias-agg (left) for the same model--dataset column.}
\label{tab:bias_agg_attr_artificial}
\end{table*}

\paragraph{Bias-agg} Table \ref{tab:bias_agg_attr_artificial} shows that \uline{\texttt{ModernBERT} has the lowest position bias} on the three datasets, in 13/18 cases and on average (.15, .31--tie, .06). Notably, proneness to position bias does not seem to be related to lexical bias: similarly to Bias-cons, the roles are reversed, with \uline{\texttt{BERT} exhibiting lowest lexical bias} in 15/18 cases and on average (.05, .04, .19). The contrast is clear-cut in \texttt{noun-det-period} (including a flipped PartSHAP), mixed in \texttt{period-comma} and clear-cut in \texttt{unique-punctuation}. Despite prior research presenting attribution methods as disagreeing, we find that they show consistency in picking the models with the least and most bias.

\paragraph{Bias-attr} The main pattern in Table \ref{tab:bias_agg_attr_artificial} is \uline{VanGrad having the highest \textit{position} Bias-attr} in 5/6 scenarios, across 3 datasets $\times$ 2 models. VanGrad's \textit{lexical} bias is less prominent, but still highest in 3/6 scenarios. There is no clear pattern of lowest bias. 

We compare Bias-attr results to Bias-agg results to determine whether the difference between methods is indicative of a difference between the methods and a no-bias-baseline. For this, we consider the highest Bias-attr method per dataset--model pair: in 6/6 cases it corresponds to the highest Bias-agg method for \textit{position} and in 4/6 cases for \textit{lexical} bias. If we look at low Bias-attr scores, this parallel pattern with Bias-agg is less clear: several methods exhibit comparably low scores. This indicates that, at least in our experiments, \uline{methods with a stronger bias compared to one another (Bias-attr) are also more biased overall compared to a uniform baseline (Bias-agg)}.

\paragraph{Overview of bias patterns in artificial experiments} Across our experiments, the following trends emerge. In terms of aggregate bias, \texttt{ModernBERT} consistently exhibits lower position bias, while \texttt{BERT} tends to have lower lexical bias, suggesting a trade-off between the two. Regarding the specific feature attribution methods, VanGrad exhibits the highest position bias in most scenarios, whereas lexical bias differences between methods are less pronounced. Bias-cons results indicate that some methods are more consistent across seeds than other, but overall, methods with stronger attribution bias tend to correspond to higher aggregate bias. This shows that both the choice of the model and the choice of the attribution can affect explanation bias, and that evaluating bias across multiple metrics is necessary to form a complete picture.

\paragraph{Sufficiency and comprehensiveness} As a sanity check for the faithfulness metrics \textit{suff} and \textit{cmp}, we assess whether they are able to detect VanGrad's high Bias-agg. Table \ref{tab:suff_comp} shows that these metrics do not reflect VanGrad having the strongest bias: while \textit{suff} = .00 would usually indicate faithful explanations, it returns .00 for other methods as well, meaning that no peculiar behaviour of VanGrad is detectable. \textit{cmp} measures the impact of the context around the top-1 explanation, and understandably is .00 as well, being equally unable to predict the random class as the full input sequence. This confirms our expectations: \textit{suff} and \textit{cmp} are not suitable to detect lexical and position bias in attribution methods for models that failed to learn the task (well). By scoring $\approx 0$, these metrics could instead be used as sanity checks for the \textit{model}, as they correctly reflect that no data artifacts (absent by design) were exploited for the prediction.

\begin{table}[h]
\scriptsize
\centering
\begin{tabularx}{\columnwidth}{l *{4}{>{\centering\arraybackslash}X}}
\toprule
& \multicolumn{2}{c}{\texttt{B}} & \multicolumn{2}{c}{\texttt{ModB}} \\
\textbf{Method} & \textit{suff} & \textit{cmp} & \textit{suff} & \textit{cmp} \\
\midrule
PartSHAP     & -.03 &  .01 & -.07 &  .01 \\
LIME         & -.02 &  .01 & -.02 &  .01 \\
VanGrad      & -.00 &  .00 &  .00 & -.00 \\
Grad\,×\,I   &  .00 &  .00 & -.04 &  .01 \\
IntGrad      & -.01 &  .00 & -.01 &  .00 \\
IntGrad\,×\,I& -.00 &  .00 & -.01 &  .00 \\
\bottomrule
\end{tabularx}
\caption{Sufficiency and comprehensiveness (\textit{suff}, \textit{cmp}) for \texttt{BERT} (\texttt{B}) and \texttt{ModernBERT} (\texttt{ModB}) on \texttt{unique-punctuation}. Multiple run averages.}
\label{tab:suff_comp}
\end{table}

\section{Experiments on Causal Relation Data}\label{sec:experiments_on_causal_relation_data}
In Section \ref{sec:experiments_on_artificial_data}, we analysed position and lexical bias on artificial data in a fully controlled experiment. While a randomised setting provides valuable insights into the sensitivity of attribution methods to exhibit bias under controlled conditions, such findings may not fully capture their behavior in real world tasks. For this reason, we complement this analysis with a semi-controlled experiment on natural language. Specifically, we fine-tune a binary classifier to detect causal relations in a text, using the same \texttt{BERT} and \texttt{ModernBERT} transformers. 

Causal relations can be expressed in many ways, both explicitly and implicitly \cite{solstadbott}. Here we use data with explicit markers, such as causal verbs ({\it cause}, {\it depend on}) and adverbial phrases ({\it because}, {\it as a result of}). These strong markers make this task particularly useful for our purpose, contrasting with the artificial dataset where there are no overt markers the models can learn from. There are two main tasks in the domain of causality detection: sentence classification (does it contain causal information?) and causal span detection (cause, marker, and effect) \cite[e.g.]{asghar2016}. We address the former task.

Position and lexical information are more intertwined in natural language tasks, with their combination forming the semantics that models can encode. One of the main differences is that we cannot expect a uniform distribution as our bias baseline: token types are not evenly distributed and class signals do not appear evenly across indices in the input. We therefore craft the task in such a way that we can study position bias, but focus on sentence position (in a text) rather than token position (in a sentence). This is based on the scenario in which multiple causal signals occur in the same input: which one is going to be responsible for the prediction? After assessing bias in both artificial and natural language experiments, we carefully compare whether any bias patterns hold for the same models and attribution methods.

\subsection{Data}
The data is adapted from \citet{veldhuisetal}, and consists of annotated texts on causal relations used for system dynamics modelling. 
Topics include climate change, putinism policy and traffic congestion \citep[e.g.]{mcfaul2020putin}.
We finetune the classifiers on alterations of these texts to i) emulate the repetition of causal signals in the same input and ii) to control for position bias compared to a baseline. For this, we first randomly sample sentence triples (900 for training, 100 for testing). We then construct input instances as permutations of these triples, with each triple resulting in 6 different orderings of the three different sentences (5400 for training, 600 for testing). Positive instances (50\% of the data) consist of concatenations of three different positive sentences (\texttt{+++}); the same logic applies to negative instances (\texttt{{-}{-}{-}}). The three sentences in \texttt{+++} represent causal occurrences at beginning-, middle- and end-of-text. We name this diagnostic dataset \texttt{causal}. The task is to predict a positive or negative class label for each \textit{instance}. We train and test on the perturbations so that the model should not learn that the dominant causal signals (each instance is expected to contain one) are tied to specific sentence positions. By only changing the order of the sentences (and not e.g.\ the syntactic structure) and keeping the tokens constant per permutation batch, we control for confounding factors that would arise from feeding different inputs. On average, positive sentences contain 24.6 words (6--60), negative 18.4 (1--57), while \texttt{+++} and \texttt{{-}{-}{-}} test instances contain 64.1 (22--128). We provide examples in \ref{sec:example_inputs}.

\subsection{Bias and Baseline Distributions}
Position bias is measured as a distribution over the three sentence positions in each of the inputs. In positive instances, each sentence will contain a signal of causality. It is reasonable to expect that the causal signal of one particular sentence will be dominant. For each set of 6 permutations, the dominant signal would appear twice in the sentence placed first, twice second and twice third. Therefore, the baseline implying absence of position bias is a uniform distribution over sentence positions.

To deepen our analysis, we differentiate between positive class attributions and negative class attributions. The former are expected to target strong causal signals, whereas the latter must deal with a signal \textit{absence}, which makes attribution intuitively more challenging \citep[e.g.][]{dela2024none}. 

\subsection{Results}
\texttt{BERT} and \texttt{ModernBERT} fine-tuned on the \texttt{causal} task achieve f1-scores $\scriptstyle>$.98, hence forming an interesting counter scenario to the artificial experiments (f1-score $\scriptstyle\sim$.50). As a first lexical insight, the targeted top-1 tokens vary across models and methods. 
For example, \texttt{BERT}'s LIME selects mostly causal verbs such as \textit{produce, increases, achieve, influence, weaken}.
\texttt{ModernBERT}'s IntGrad selects a mix of different lexical elements, such as causal verbs (\textit{made, change, developed}), but also punctuation, determiners and topic-related words (\textit{temperature, climate, regime, emissions}). The latter implausible explanations increase the relevance of investigating \textit{where} in the input they occur. Moreover, methods attribute more function words and punctuation in negative examples, making up for the absence of causal signals. In this section, we report the sentence position biases from these models. 

\begin{figure}[h]
  \centering
  \begin{subfigure}{.49\textwidth}
    \includegraphics[width=\textwidth]{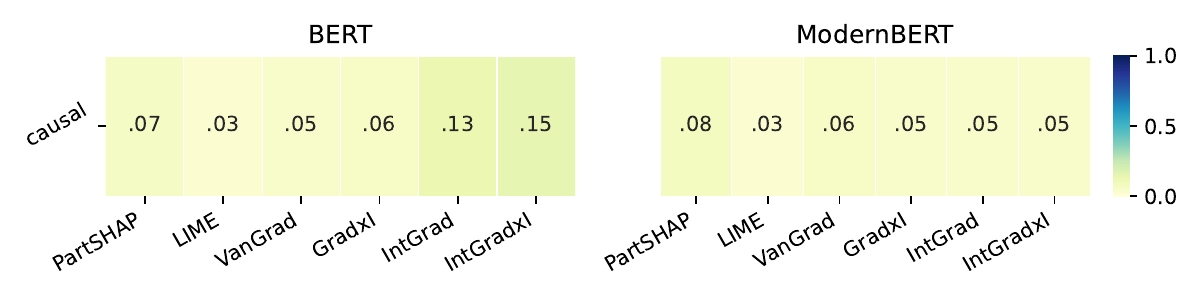}
    \caption{Positive class Bias-cons for sentence position.}
    \label{fig:_2plots_sentence_hits_JS_interseed_CAUSAL_POS}
  \end{subfigure}
  
  \vspace{0.1cm} 
  
  \begin{subfigure}{.49\textwidth}
    \includegraphics[width=\textwidth]{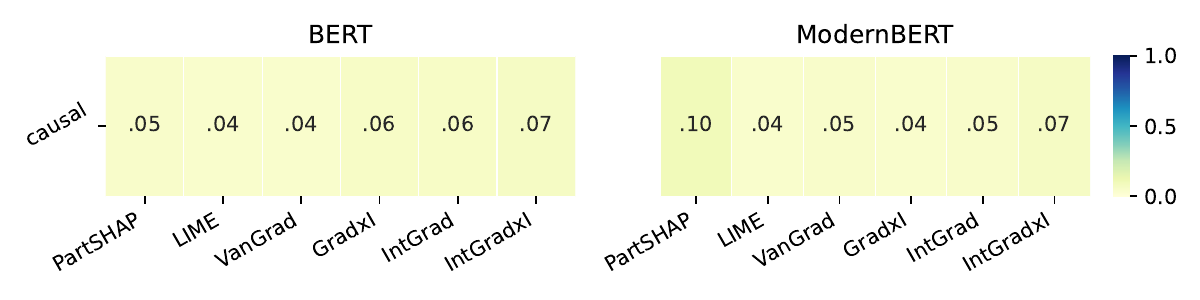}
    \caption{Negative class Bias-cons for sentence position.}
    \label{fig:_2plots_sentence_hits_JS_interseed_CAUSAL_NEG}
  \end{subfigure}

  \caption{Bias-cons: inter-seed consistency on 10 different runs. Low values (light BG) = greater consistency.}
  \label{fig:_2plots_sentence_hits_JS_interseed_CAUSAL}
\end{figure}

\begin{figure*}[h]
  \centering
  \begin{subfigure}{0.48\textwidth}
    \includegraphics[width=\textwidth]{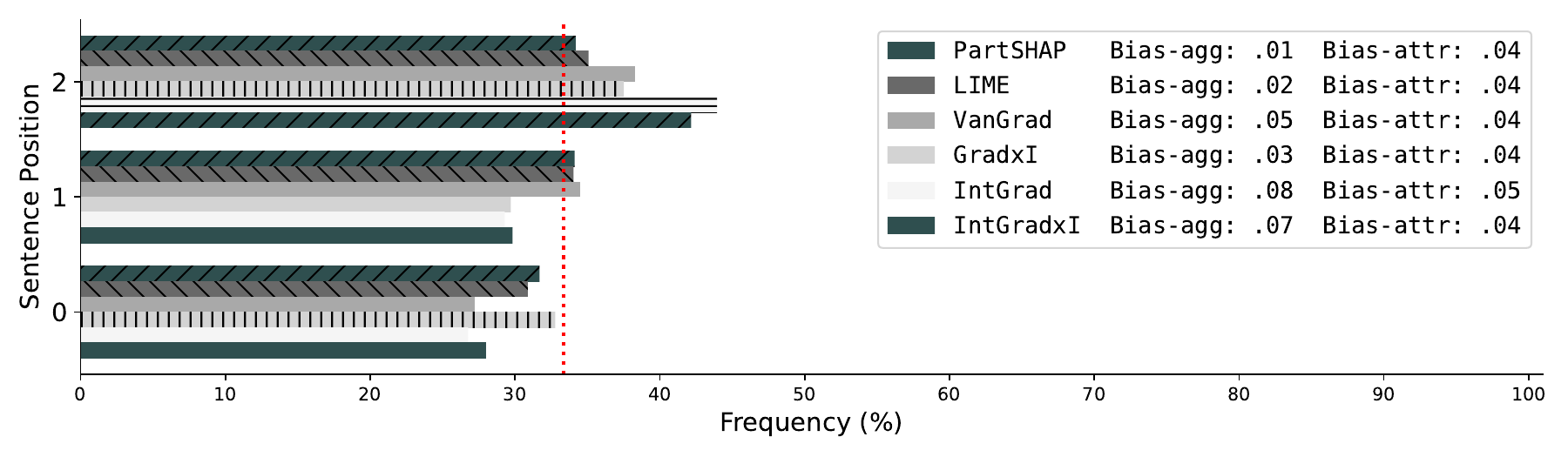}
    \caption{\texttt{BERT}, positive class.}
    \label{fig:_bert_sentencehits_barplot_exp1_onlypositive}
  \end{subfigure}
  \begin{subfigure}{0.48\textwidth}
    \includegraphics[width=\textwidth]{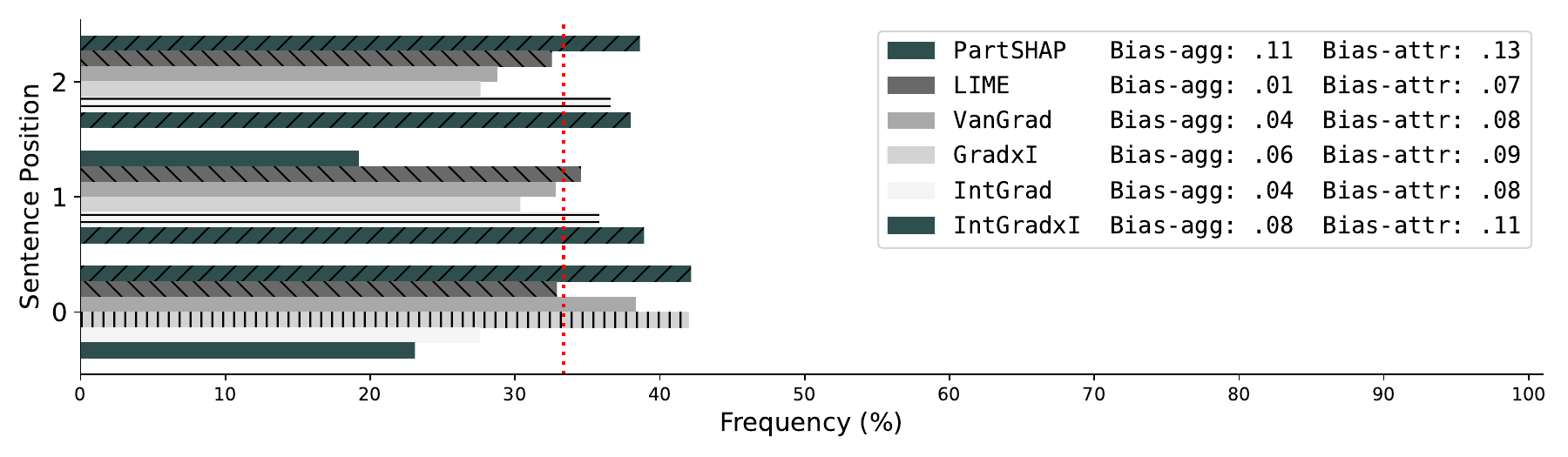}
    \caption{\texttt{ModernBERT}, positive class.}
    \label{fig:_modernbert_sentencehits_barplot_exp1_onlypositive}
  \end{subfigure}
  
  \vspace{0.1cm} 
  
  \begin{subfigure}{0.48\textwidth}
    \includegraphics[width=\textwidth]{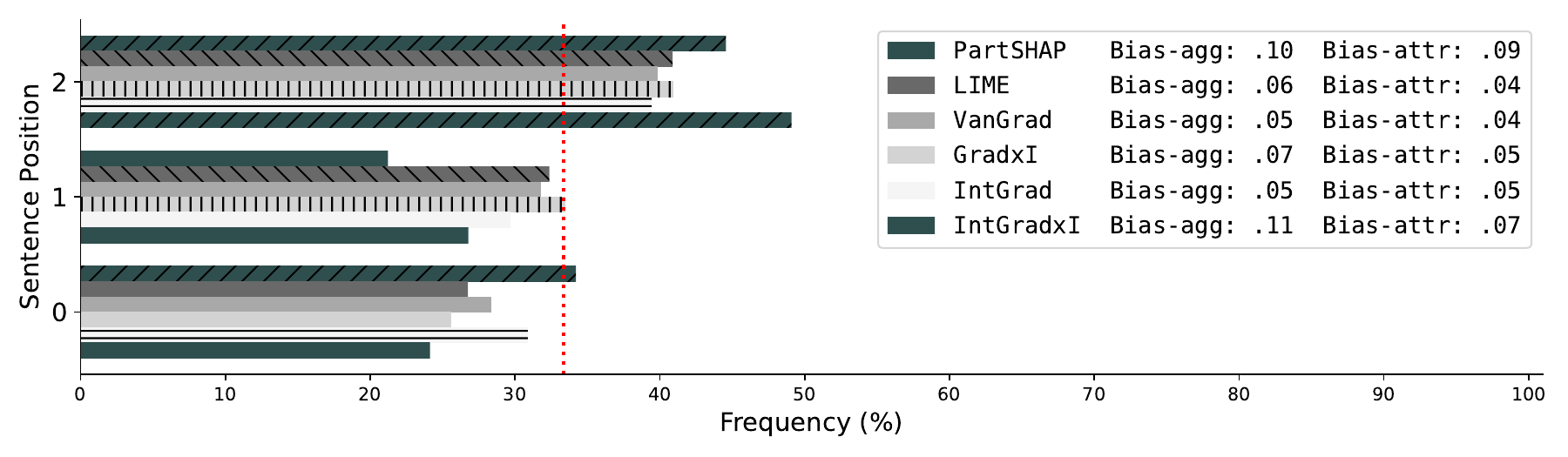}
    \caption{\texttt{BERT}, negative class.}
    \label{fig:_bert_sentencehits_barplot_exp1_onlynegative}
  \end{subfigure}
  \begin{subfigure}{0.48\textwidth}
    \includegraphics[width=\textwidth]{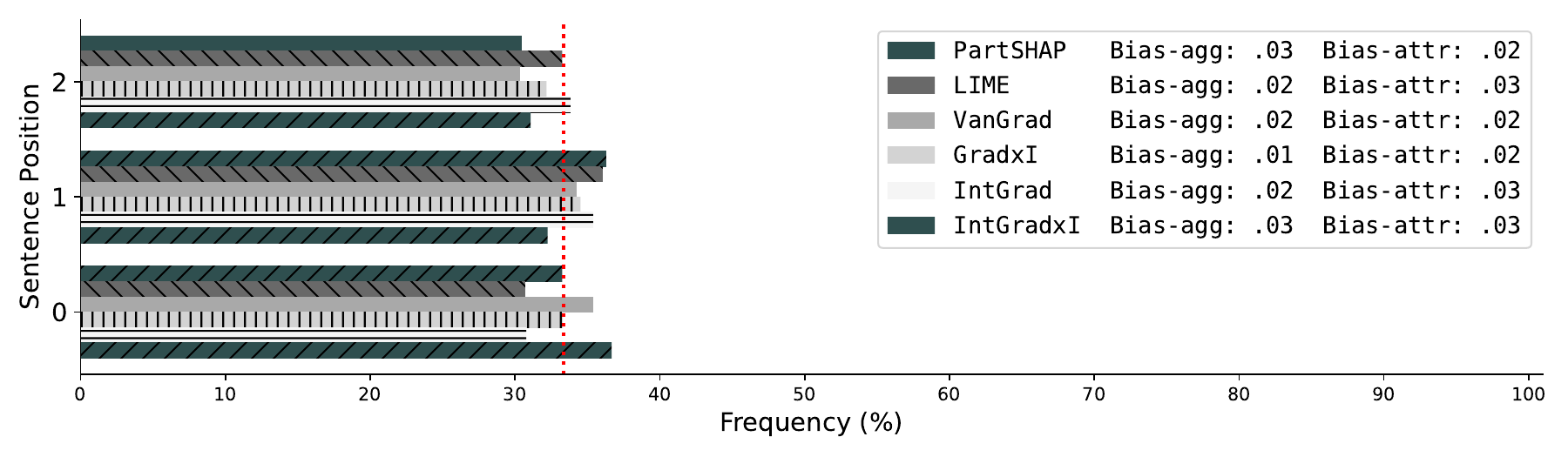}
    \caption{\texttt{ModernBERT}, negative class.}
    \label{fig:_modernbert_sentencehits_barplot_exp1_onlynegative}
  \end{subfigure}

  \caption{Sentence position Bias-agg (relative to input triple) in causal relation detection. Top row: positive class (\texttt{+++}, causal); bottom row: negative class (\texttt{{-}{-}{-}}, non-causal). Vertical red line indicates chance.}
  \label{fig:_all_sentencehits_barplot_exp1}
\end{figure*}

\paragraph{Bias-cons} Figure \ref{fig:_2plots_sentence_hits_JS_interseed_CAUSAL} shows that attribution methods yield a relatively consistent bias across sentence positions (.03--.15 across models). This suggests that \uline{a dominant causal signal tends to be identified consistently across seeds irrespective of position}. Low scores are in part due to a relatively low number of sentence positions (3), facilitating the task by conflating token attributions into sentence attributions. While LIME and IntGrad were the most consistent on the artificial data, only LIME is also the most consistent on \texttt{causal}, suggesting that \uline{LIME is only lightly affected by the seed}. 
Different from the artificial setting, \uline{\texttt{ModernBERT} is not predominantly less biased than \texttt{BERT}}, despite \texttt{BERT}'s higher bias on IntGrad and IntGrad\,×\,I (.13 and.15). 

\paragraph{Bias-agg and Bias-attr} For \texttt{BERT}, the attributions point to the dominant (non-)causal signal appearing with a slight bias in the last sentence of each \texttt{+++} or \texttt{{-}{-}{-}} instance (Figure \ref{fig:_all_sentencehits_barplot_exp1}). This suggests that \uline{\texttt{BERT} is most biased towards late information in the input}, especially on \textit{negative} examples, which is in line with the results on artificial data (Figures \ref{fig:_barplot_noun-det-period}, \ref{fig:_barplot_period-comma}, \ref{fig:_barplot_unique-punctuation}, Appendix \ref{appendix:A}). 

\uline{\texttt{ModernBERT} is overall less biased than \texttt{BERT}}, which also is in line with the artificial setting (Table \ref{tab:bias_agg_attr_artificial}). This is especially visible on \textit{negative} examples (Bias-agg .01--.03), with methods being aligned on this bias pattern (Bias-attr .02--.03). Interestingly enough, Bias-agg is higher and more variable on \texttt{ModernBERT} $\times$ positive class, with PartSHAP (.11) and LIME (.01) standing in sharp contrast. 

In line with the artificial data, we find a similar relation between Bias-agg and Bias-attr: \uline{the method with the highest Bias-attr also exhibits the highest Bias-agg} for both models on positive class (IntGrad--Fig. \ref{fig:_bert_sentencehits_barplot_exp1_onlypositive}; PartSHAP--Fig. \ref{fig:_modernbert_sentencehits_barplot_exp1_onlynegative}), and on the top two methods on \texttt{BERT} $\times$ negative class (PartSHAP and IntGrad\,×\,I--Fig. \ref{fig:_bert_sentencehits_barplot_exp1_onlynegative}). In Fig. \ref{fig:_modernbert_sentencehits_barplot_exp1_onlynegative}, a pattern is hardly found, but biases are already low.

\paragraph{Overview of bias patterns in causal experiments} Across the causal relation detection task, the following trends emerge. First, Bias-cons is generally low, indicating that the position of dominant signals is generally identified consistently across seeds. Explanations generated for \texttt{BERT} tend to focus more on later sentence positions, especially for negative examples, whereas \texttt{ModernBERT} exhibits lower and more uniform bias overall, consistent with the artificial experiments. Regarding the specific feature attribution methods, LIME has the highest consistency across different random seeds, while other methods' (relative) consistency scores appear to be more model-dependent. As in the artificial setting, methods with the highest aggregate bias are inclined to also exhibiting the highest attribution bias, carefully suggesting that an anomalous explainer might be more likely to generate unreliable explanations. 

\section{Conclusion and Future Directions}
We proposed a bias evaluation framework on both artificial and natural language data, finding systematic preferences for lexical and positional information. We investigated how lexical and position bias relate and whether a recent BERT transformer would be less biased than its traditional version. We conclude that there is a trade-off between the two types of bias and that \texttt{ModernBERT} is not necessarily less biased than \texttt{BERT}. We also investigated whether evidence from anomalous explainers is more likely to be incorrect. To this end, we found repeated empirical evidence that methods differing most from other methods are also the most biased. Since our settings controlled for lexical and position confounders, this bias is expected to stem from the method itself and not from the backbone model.

Our results show the importance of testing attribution bias from different angles. Besides our insights on negative class examples (lacking strong signals), and the seeming inability of classical faithfulness metrics to recognise unfaithful attribution methods, we encourage further research on models that only partially learn a task. We expect that these metrics would also be partially reliable. Another direction is to explore the attribution preference for contextual tokens \textit{beyond} the strongest signal (e.g.\ top-5$\setminus$top-1). Given that weaker signals would appear among contextual tokens, the disagreement between attribution would likely be higher, leading to biases being accentuated in such a scenario.

\section*{Limitations}
We report some limitations of our work. The first concerns the sample of feature attribution methods that were adopted in our analyses; while it includes methods from both gradient-based and perturbation-based families, they are not directly representative of other methods outside this sample. The risk exists that our results are inadequately generalised for other explainable AI methods, even unrelated to feature attribution. Similarly, we crafted multiple data variations for the artificial setting and observed their common traits as a showcase of our proposed framework; however, the possibilities for non-natural lexical combinations are potentially infinite, and may be refined with more precise hypotheses in future work. Again, the risk exists that the results from these specific datasets are generalised over \textit{any} controlled setting with random token sequences. Finally, controlled experimental settings favour phenomenon isolation but also limit the degrees of comparison with less controlled settings; this applies to the natural language task, which had to rely on different baseline assumptions (non-uniform lexical preferences) and different language units (sentences versus tokens) than the controlled, artificial setting.

\section*{Acknowledgments}
This research is funded by the Dutch National Science Organisation (NWO) through project code NWA.1292.19.399 -- \textit{InDeep: Interpreting Deep Learning Models for Text and Sound}. We would also like to thank TNO and the APPL.AI programme for their support of this project, and Antske Fokkens and Lisa Beinborn for the feedback. Remaining errors are our own.
\bibliography{custom}

\appendix
\section{Appendix}\label{appendix:A}


\subsection{Example Inputs}\label{sec:example_inputs}
We provide a positive (\texttt{+++}) and a negative (\texttt{{-}{-}{-}}) input instance from the \texttt{causal} dataset on which predictions and attributions were computed.





\begin{tcolorbox}[examplebox]
\textbf{\texttt{+++}}

\begin{tcolorbox}[examplebox]
Still more believe that stabilizing emissions near current rates would stabilize the climate, when in fact emissions would continue to exceed removal, increasing GHG concentrations and radiative forcing. 
\end{tcolorbox}

\begin{tcolorbox}[examplebox]
The amount of energy radiated back into space depends on the composition of the atmosphere.
\end{tcolorbox}

\begin{tcolorbox}[examplebox]
People tend not to enjoy longer travel times, thus, once travel times increase a pressure will build up to reduce it.
\end{tcolorbox}
\end{tcolorbox}





\begin{tcolorbox}[examplebox]
\textbf{\texttt{{-}{-}{-}}} 

\begin{tcolorbox}[examplebox]
Both of these explanations allow no role for the agency of individual leaders and their ideas in the analysis. 
\end{tcolorbox}

\begin{tcolorbox}[examplebox]
At the Munich Security Conference in 2016, Prime Minister Dmitry Medvedev referenced the Cuban missile crisis as a similar moment in bilateral tensions: Speaking bluntly, we are rapidly rolling into a period of a new cold war. 
\end{tcolorbox}

\begin{tcolorbox}[examplebox]
The new Syrian generals empowered by a negotiated transition would likely have remained loyal to Moscow.
\end{tcolorbox}
\end{tcolorbox}

\subsection{Hardware and Runtimes}
The runtimes for fine-tuning the classifiers and generating the explanations are reported in Table \ref{tab:runtimes}. All models were fine-tuned on NVIDIA GeForce RTX 2080 Ti (11264 MiB), which was also used to generate the explanations. We used a batch size of 8 and we set the learning rate to 5{e}-6 with a decay factor of 1{e}-2.

\begin{table}[ht]
\scriptsize
\centering
\begin{tabularx}{\columnwidth}{l *{4}{>{\centering\arraybackslash}X}}
\toprule
\textbf{Model} & \textbf{Size} & \textit{fine-tuning} & \textit{explaining} \\
\midrule
\multicolumn{4}{c}{\textbf{artificial}} \\
\midrule
\texttt{BERT}       & 110M & 2 min  & $\sim$2--5 sec \\
\texttt{ModernBERT} & 149M & 3 min  & $\sim$2--5 sec \\
\midrule
\multicolumn{4}{c}{\texttt{\textbf{causal}}} \\
\midrule
\texttt{BERT}       & 110M & 6 min  & $\sim$7--9 sec \\
\texttt{ModernBERT} & 149M & 8 min  & $\sim$16--18 sec \\
\bottomrule
\end{tabularx}
\caption{Model fine-tuning times (5 epochs) on artificial and \texttt{causal} data along with explanation generation times per instance (6 default methods in Ferret \citep{attanasio-etal-2023-ferret} with default parameters).}
\label{tab:runtimes}
\end{table}


\subsection{Data Artifacts}
We use the existing dataset by \citet{veldhuisetal}. It was obtained on request to the authors, who agreed to publish the data splits used in our experiments on our repository. We advice to contact the authors for re-use of the data, intended for research purposes. The data was manually checked for offensive content or personal information, which were not encountered.

\subsection{Complementary Visualisations}\label{sec:appendix_barplots}
\begin{figure*}[htbp]
  \centering
  \begin{subfigure}[b]{0.49\textwidth}
    \includegraphics[width=\textwidth]{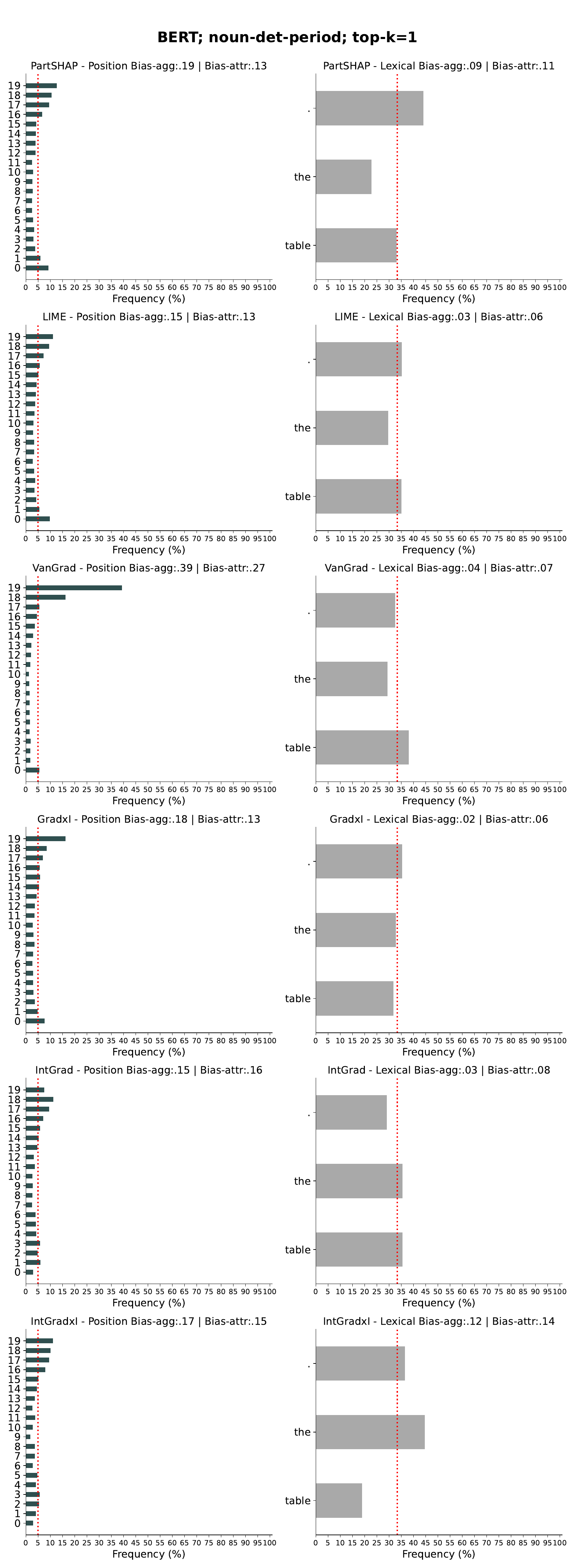}
    \caption{\texttt{BERT}}
  \end{subfigure}
  \begin{subfigure}[b]{0.49\textwidth}
    \includegraphics[width=\textwidth]{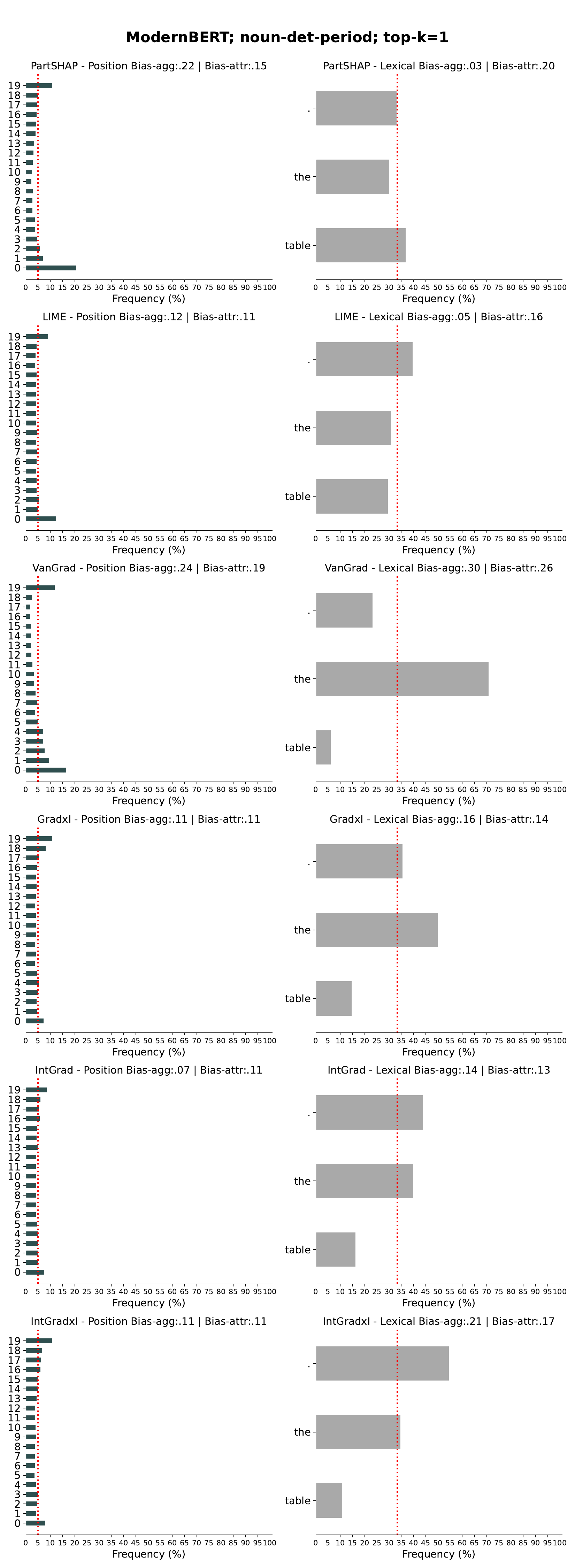}
    \caption{\texttt{ModernBERT}}
  \end{subfigure}
  \caption{\texttt{noun-det-period} biases. Different attribution methods show different top-1 preferences for lexical elements and their position in the sentence. Left: position frequency distribution. Right: lexical frequency distribution. Vertical red line indicates chance.}
  \label{fig:_barplot_noun-det-period}
\end{figure*}

\begin{figure*}[htbp]
  \centering
  \begin{subfigure}[b]{0.49\textwidth}
    \includegraphics[width=\textwidth]{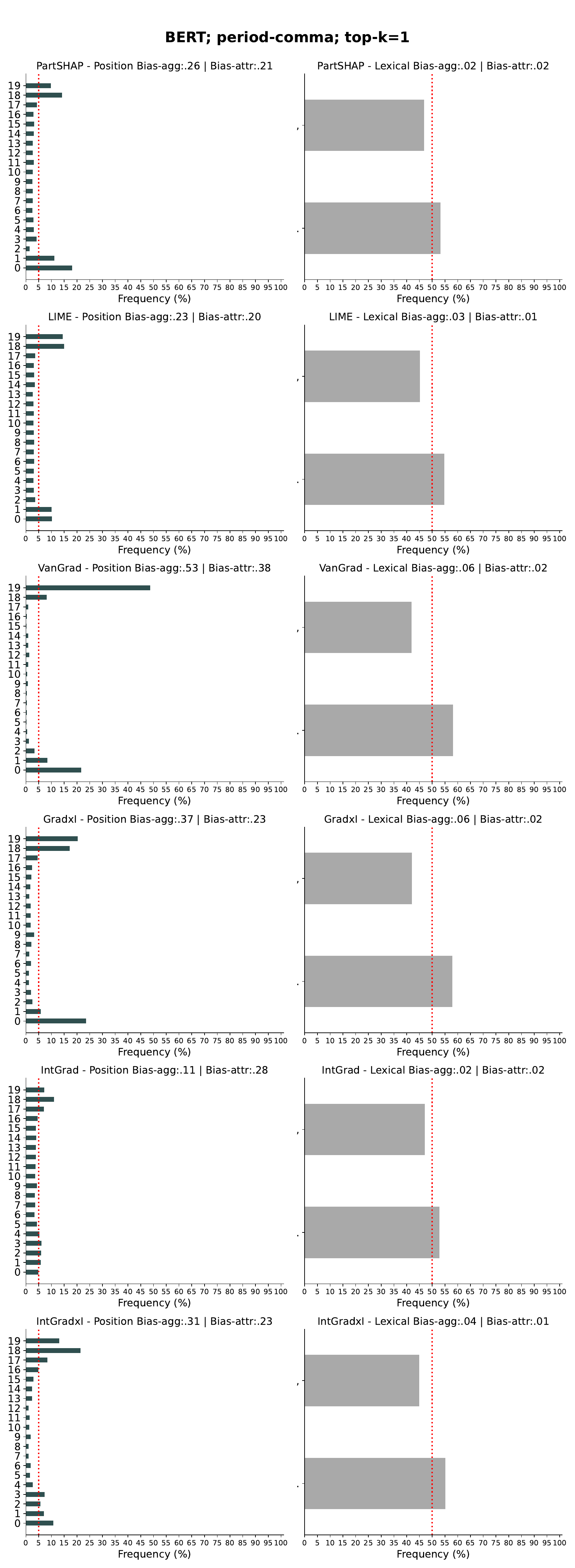}
    \caption{\texttt{BERT}}
  \end{subfigure}
  \begin{subfigure}[b]{0.49\textwidth}
    \includegraphics[width=\textwidth]{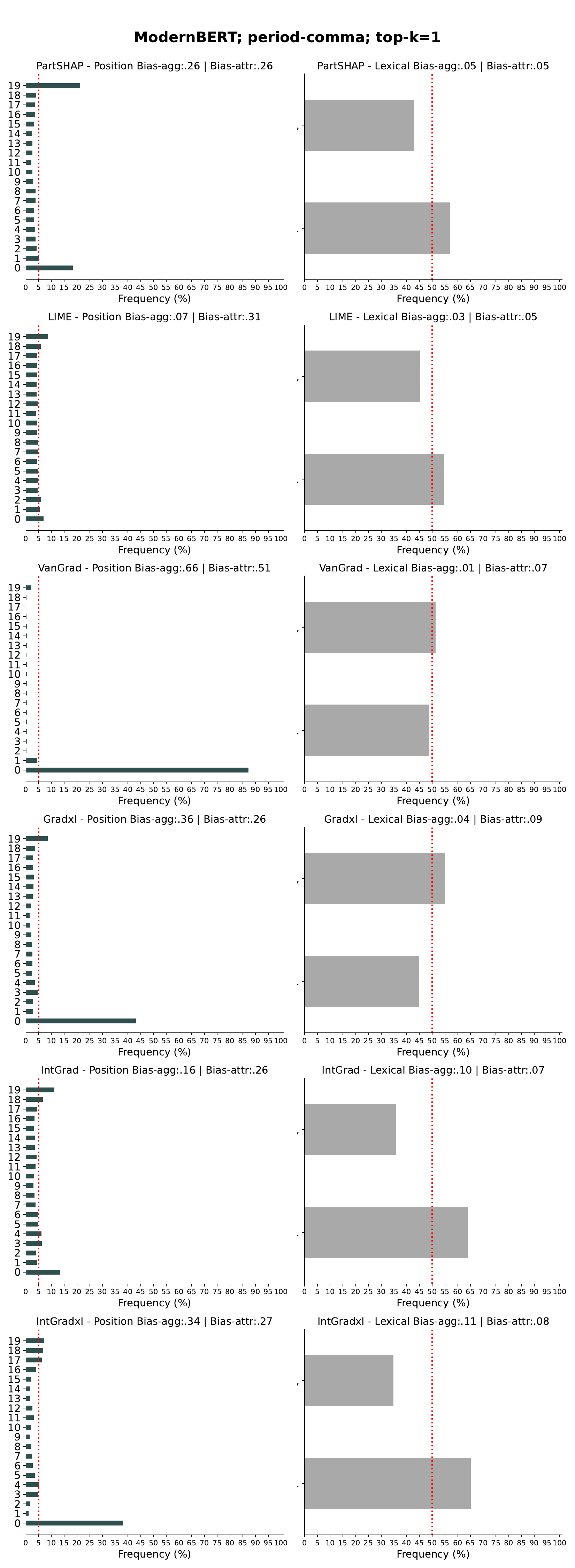}
    \caption{\texttt{ModernBERT}}
  \end{subfigure}
  \caption{\texttt{period-comma} biases. Different attribution methods show different top-1 preferences for lexical elements and their position in the sentence. Left: position frequency distribution. Right: lexical frequency distribution. Vertical red line indicates chance.}
  \label{fig:_barplot_period-comma}
\end{figure*}

\begin{figure*}[htbp]
  \centering
  \begin{subfigure}[b]{0.49\textwidth}
    \includegraphics[width=\textwidth]{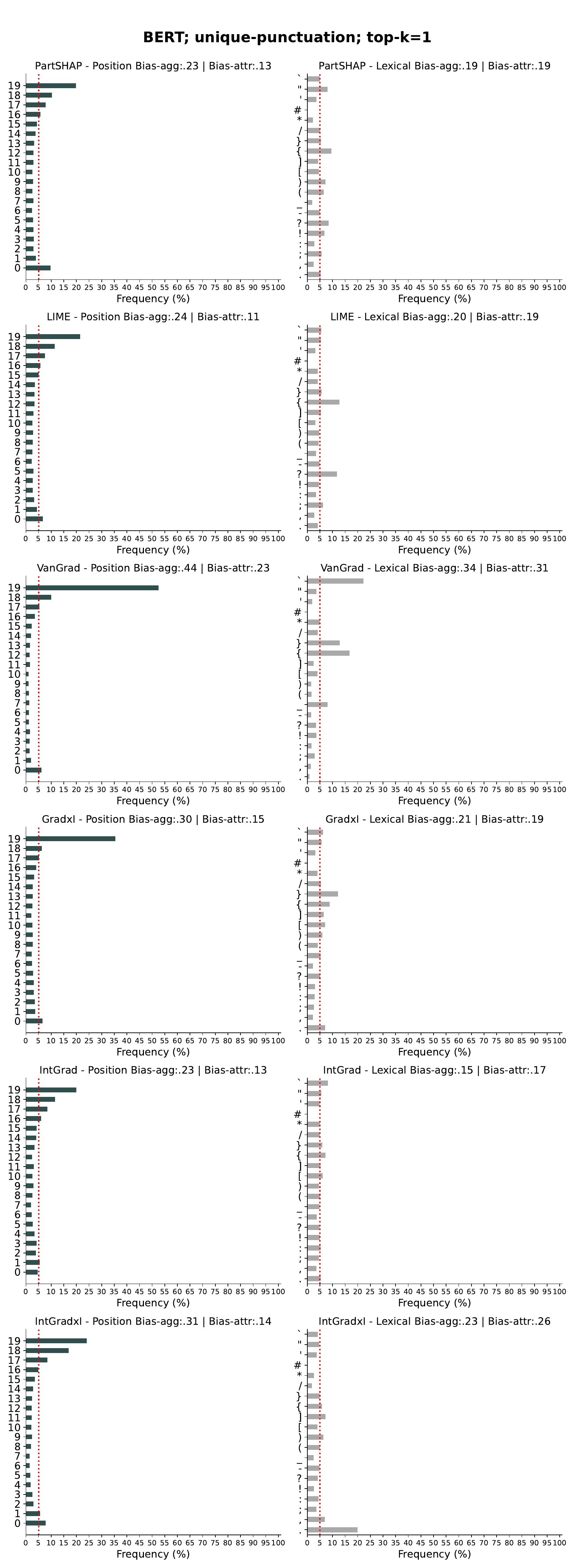}
    \caption{\texttt{BERT}}
  \end{subfigure}
  \begin{subfigure}[b]{0.49\textwidth}
    \includegraphics[width=\textwidth]{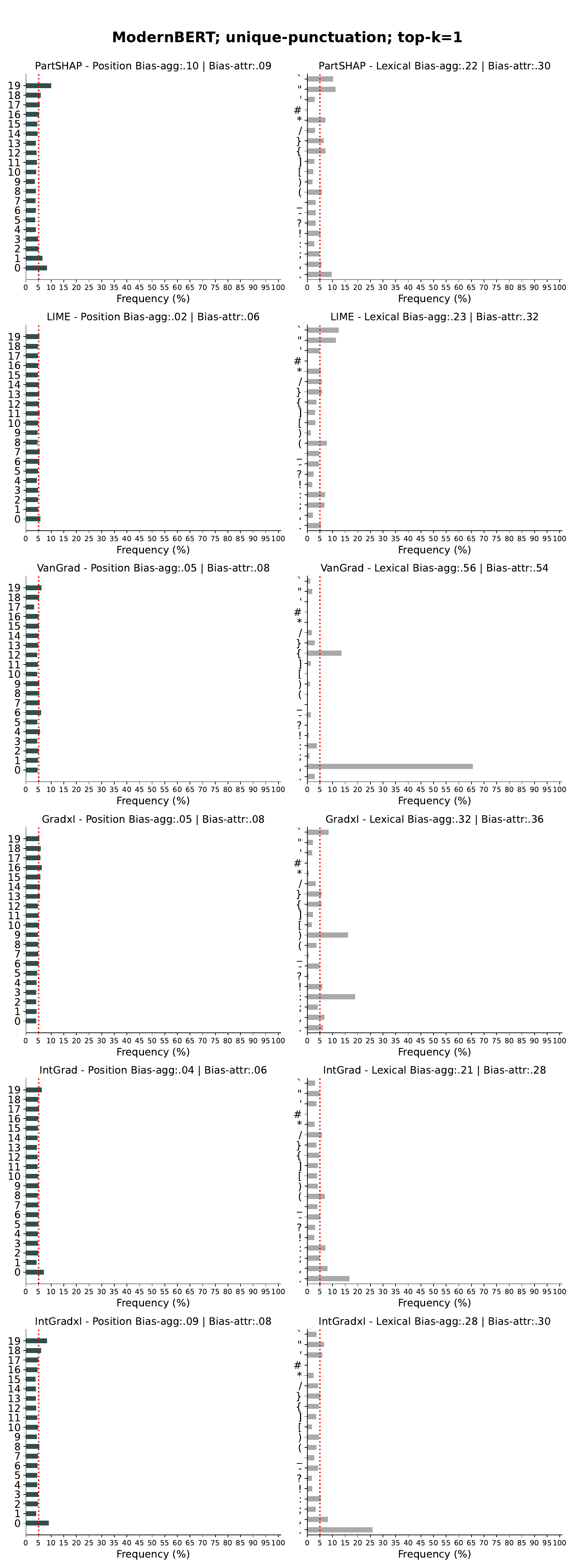}
    \caption{\texttt{ModernBERT}}
  \end{subfigure}
  \caption{\texttt{unique-punctuation} biases. Different attribution methods show different top-1 preferences for lexical elements and their position in the sentence. Left: position frequency distribution. Right: lexical frequency distribution. Vertical red line indicates chance.}
  \label{fig:_barplot_unique-punctuation}
\end{figure*}

We provide the full overview of bias distributions for all attribution methods and model types in Figure \ref{fig:_barplot_noun-det-period}, Figure \ref{fig:_barplot_period-comma} and Figure \ref{fig:_barplot_unique-punctuation} (one for each artificial dataset). These form the foundation for the bias computations reported in §\ref{sec:artificial_results}.

\end{document}